\documentclass[conference]{IEEEtran}
\IEEEoverridecommandlockouts
\usepackage{listings}
\usepackage{float}
\usepackage[normalem]{ulem}
\usepackage[utf8]{inputenc}
\usepackage{bbding}
\usepackage[T1]{fontenc}
\usepackage[textsize=scriptsize, disable]{todonotes} 
\usepackage{lipsum}
\usepackage{algorithm}
\usepackage{algpseudocode}
\usepackage{url}
\usepackage{blindtext}
\usepackage[separate-uncertainty = true]{siunitx}
\usepackage{color,soul}
\usepackage{svg}
\usepackage{booktabs}
\usepackage{amsmath} 
\usepackage{amsfonts}
\usepackage{amssymb}  
\usepackage[nolist]{acronym}
\usepackage{courier}
\usepackage{xcolor}

\def\BibTeX{{\rm B\kern-.05em{\sc i\kern-.025em b}\kern-.08em
    T\kern-.1667em\lower.7ex\hbox{E}\kern-.125emX}}
\begin{document}

\title{\huge Flexible and Adaptive Manufacturing by Complementing Knowledge Representation,  Reasoning and Planning with Reinforcement Learning \\
}

\author{\IEEEauthorblockN{1\textsuperscript{st} Matthias Mayr}
\IEEEauthorblockA{\textit{Department of Computer Science} \\
\textit{Lund University}\\
matthias.mayr@cs.lth.se}
\and
\IEEEauthorblockN{2\textsuperscript{nd} Faseeh Ahmad}
\IEEEauthorblockA{\textit{Department of Computer Science} \\
\textit{Lund University}\\
faseeh.ahmad@cs.lth.se}
\and
\IEEEauthorblockN{3\textsuperscript{rd} Volker Krueger}
\IEEEauthorblockA{\textit{Department of Computer Science} \\
\textit{Lund University}\\
volker.krueger@cs.lth.se}
}

\maketitle

\section{Introduction}
Shifting from mass manufacturing towards greater customizations and smaller batch sizes requires specialized architectures and solutions capable of effectively managing this change. A system designed for this transition must not only exhibit a high degree of flexibility but it must also be able to cope with the added complexity introduced by the integration of robotic systems such as manipulators and their different hardware. Finally, it is desirable that even for contact-rich manufacturing tasks, the system should be able to learn from data and improve the execution.
In the light of these objectives, we present our work that leverages task-level planning and reasoning and combines it with reinforcement learning. The former provides a structure for a task, while the latter allows to learn and improve the execution from interaction data. This synergy creates a comprehensive pipeline that translates high-level planning goals, often originating from a manufacturing execution system, into tangible improvements in execution within a multi-objective setting on the robot system.

In the following sections, we provide an overview of the skill-based system \textit{SkiROS2} and its planning and reasoning capabilities. We introduce the integration of multi-objective reinforcement learning and our approach to address reinforcement learning for variations of tasks.

\section{SkiROS2 - A Skill-based Robot System}
\begin{figure}[tpb]
	{
		\setlength{\fboxrule}{0pt}
		\framebox{\parbox{3in}{
        \centering
		\includegraphics[width=0.75\columnwidth]{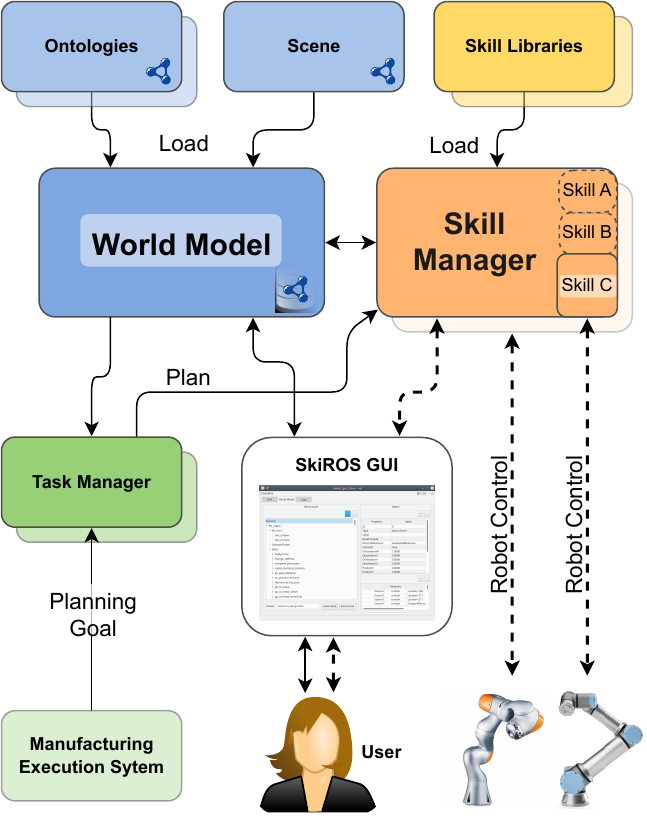}}
        }
	}
	\caption{An outline of the SkiROS2 architecture. The world model stores the knowledge about the relations, environment and the skills. The skill manager loads and executes the skills. Dashed lines show control flows and solid lines information flows. Shaded blocks indicate possible multiple instances.}
	\label{fig:architecture}
\end{figure}

\textit{SkiROS2}~\cite{mayr23iros} is a skill-based robot control platform for the Robot Operating System (ROS). Its primary application lies within semi-structured workspaces where an initial understanding of the world exists, but requires ongoing correction and adaptation throughout the execution process. An overview of the architecture is shown in Figure~\ref{fig:architecture}.

A core component of \textit{SkiROS2} is its world model, which employs a Resource Description Framework (RDF) graph as the foundation for knowledge representation. Ontologies define the known concepts, properties and relations. Additionally, a scene holds the concrete instances (vocabulary) of these concepts. This knowledge can additionally be used to catch bugs in skills~\cite{rizwan2023ezskiros}. One or more skill managers are responsible for not only retrieving skills from skill libraries but also executing them on a robot system. Lastly, the task manager plays a pivotal role in receiving planning goals and generating plans, taking into account the current state of the world model.

\textit{SkiROS2} has a broad definition of a \textit{skill}. This allows to integrate a diverse range of existing solutions, ranging from motion planning to deep-learning-based vision. A \textit{skill description} descibes the skill on a semantic level, specifying its parameters and the pre-, hold-, and post-conditions. Following the extended behavior tree (BT) formalism~\cite{rovida172iicirsi}, these conditions can be directly employed to construct a planning domain. \textit{Skill implementations} can implement a skill description. This implementation can take the form of either a semantically atomic \textit{primitive skill} or a \textit{compound skill}. The latter allows the combination of multiple other skills within a BT.
A comparison with other skill-based systems can be found in~\cite{mayr23iros}.

\section{Planning, Knowledge Representation and Reasoning}
\textit{SkiROS2} has a well developed toolset to support flexible manufacturing. The task manager can receive planning goals in the widely supported Planning Domain Definition Language (PDDL) and automatically generate a domain and problem description based on the loaded skills and the current state of the world model.

The knowledge representation within the world model, utilizing an RDF graph, enables an explicit representation of available knowledge that can be easily and automatically generated and modified. This architecture explicitly separates the task knowledge from the skill implementation, promoting adaptability across various robot hardware and tasks~\cite{mayr2023using}. In this light, we use behavior trees and motion generators (BTMG) policy representation for skill implementation to solve tasks~\cite{rovida18btmg}. Finally, the relations defined by the ontologies facilitate reasoning capabilities, such as spatial reasoning~\cite{rovida18btmg}, in the parameterization and execution of tasks.

\section{Reinforcement Learning}
\begin{figure}[tpb]
	{
		\setlength{\fboxrule}{0pt}
		\framebox{\parbox{3in}{
		\includegraphics[width=0.92\columnwidth]{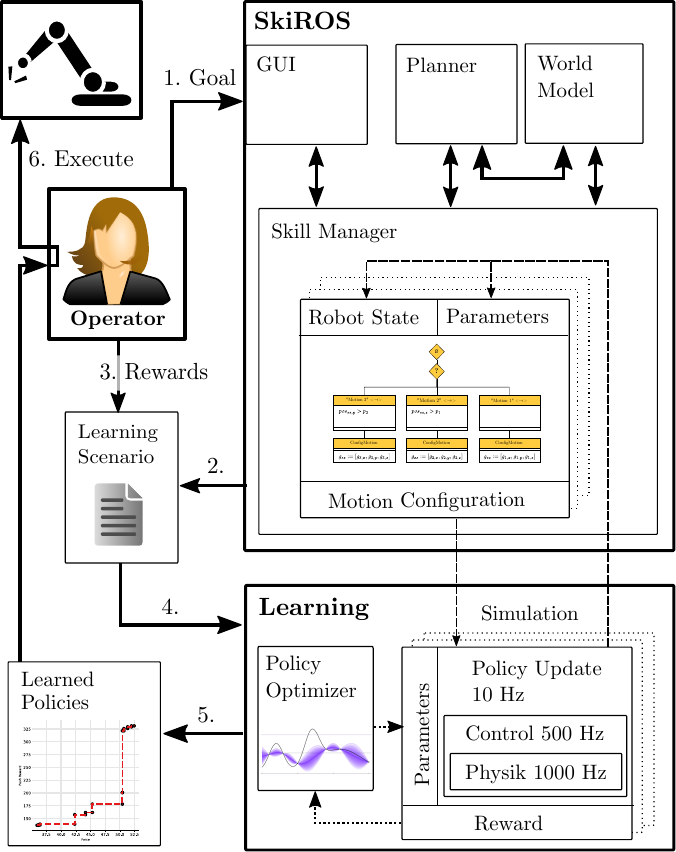}}}
	}
	\caption{The architecture of the system that depicts the pipeline: (1) The operator enters the goal state; (2) a learning scenario for the plan is created; (3) rewards and hyperparameters are specified; (4) learning is conducted using the skills and the information in the world model; (5) after policy learning, the operator can choose which policies to execute on the real system (6).}
	\label{fig:system}
	\vspace{-1em}
\end{figure}

A recent line of research has explored to learn parameters of skills~\cite{mayr21iros, mayr2022combining, mayr2022skill, mayr22priors}. Specifically, in~\cite{mayr2022combining, mayr2022skill} the integration of reinforcement learning (RL) with robot manipulators into the planning and reasoning pipeline is introduced. As shown in Figure~\ref{fig:system}, the operator is actively involved in the creation of the RL scenario and the final selection of the policy. Formulating RL problems is challenging and is specifically difficult in the industrial domain that has many requirements, such as safety constraints. With this learning formulation allowing for multi-objective RL, it actively acknowledges that there is not only a single performance indicator. This eases the learning scenario design and gives the operator a selection of policies along the Pareto frontier to choose from. In~\cite{mayr22priors}, an integration of parameter priors for the optimum is introduced. This allows an operator or another source of experience to specify supposedly optimal regions in the parameter space to speed up the search. The methods have been shown to allow for learning contact-rich tasks with compliant control~\cite{mayr2022c++} either directly on the real system~\cite{mayr21iros, mayr22priors} or in simulation and transferring to the real system~\cite{mayr21iros, mayr2022skill, mayr22priors}.

\section{Learning for Task Variations}
As stated in the introduction, adaptation to different task configurations is important. While \textit{SkiROS2} as a skill-based system fully supports to adapt to task variations, depending on the influence of the learned skill parameters, the policies learned in~\cite{mayr21iros, mayr2022combining, mayr2022skill, mayr22priors} can only transfer for some changes in the tasks. This has been identified in~\cite{ahmad2022generalizing} and addressed in~\cite{ahmad2023learning} as a supervised learning problem. In short, in~\cite{ahmad2023learning} a set of BTMG parameters are learned for random task variations using the previously introduced learning pipeline. The results of all learning episodes in all variations are used to train a model that combines Gaussian Processes and a weighted support vector machine classifier. This model learns both the reward and feasibility functions based on skill and task variation parameters. Rewards represent the task performance, while feasibility predicts successful task execution with specific parameter combinations. We further optimize the model to derive a feasible policy with maximum reward. When a parameterization for an unseen task variation is requested, a combination of the planning and reasoning capabilites of \textit{SkiROS2} and the inference of the learned model is used to provide a full parameterization. The results and comparisons with baseline methods are presented in~\cite{ahmad2023learning}.

\section{Conclusions}
Modern manufacturing needs solutions that are flexible and adaptive. We have outlined a way to combine task-level planning, knowledge representation and reasoning with reinforcement learning techniques. Not only does this combination allow for easy adaption of tasks, we have also shown how robot systems can learn and improve the task execution. The formulation as a multi-objective learning problem eases the learning problem definition and allows the operator to choose a suitable policy. The possibility to integrate user priors into the search does not only allow an experienced operator to state promissing regions, but has also shown to be able to speed up learning and make it safer.

Finally, the recent introduction of a machine learning model that can appropriately address task variations for the learned skill parameters makes it more fit for the use in future factories.
\newpage
\bibliography{root}
\bibliographystyle{bib/IEEEtran}

\end{document}